%% file: emnlp2023.tex
\pdfoutput=1
\documentclass[11pt]{article}
\usepackage{booktabs}
\usepackage[table]{xcolor}
\usepackage{cellspace}
\usepackage[]{EMNLP2023}

\usepackage{times}
\usepackage{latexsym}
\usepackage{graphicx}
\usepackage{subcaption}
\usepackage{amsmath}
\usepackage{amssymb}
\usepackage{multirow}
\usepackage{colortbl}
\usepackage{color}

\usepackage[T1]{fontenc}
\usepackage[utf8]{inputenc}

\usepackage{microtype}

\usepackage{inconsolata}

\title{CLEAN–EVAL: Clean Evaluation on Contaminated Large Language Models}

\author{
 Wenhong Zhu$^1$  \thanks{\ \ Partial work was done when Wenhong Zhu was interning at FaceMind Corporation. }  \quad Hongkun Hao$^1$ \quad Zhiwei He$^{1}$ \quad Yunze Song$^{2}$ \\ 
 \bf Yumeng Zhang$^3$ \quad Hanxu Hu$^4$ \quad  Yiran Wei$^2$ \quad Rui Wang$^{1}\footnotemark[2]$ \quad Hongyuan Lu$^{2}$\thanks{\ \ Hongyuan Lu and Rui Wang are corresponding authors.}\\
$^1$Shanghai Jiao Tong University \quad $^2$FaceMind Corporation \\ $^3$Tsinghua University \quad $^4$University of Edinburgh \\
{ \normalsize \{zwhong714, haohongkun, zwhe.cs, wangrui12\}@sjtu.edu.cn} \\
{ \normalsize hongyuanlu@outlook.com} \\
}

\begin{document}
\maketitle

\input{abstract.tex}
\input{intro.tex}
\input{re.tex}

\input{method.tex}

\input{exp}
\input{ana}

\input{case_study}
\input{conclusion}
\section*{Limitations}

\paragraph{Datasets.} This paper focuses on two mainstream models. Without knowledge regarding their training data, our selected benchmark, mimicking the no-contamination setting, likely overlaps with their existing training data. Consequently, performance testing on these benchmarks could yield inflated performance metrics. Moreover, we sampled approximately 300 instances for each benchmark due to resource constraints. However, despite this limited number, randomness in sampling aims to ensure these instances represent the entire dataset.

\paragraph{Fine-tuning.} Given the extensive collection of benchmarks, conducting exhaustive fine-tuning to maximize model performance becomes impractical. Instead, we fine-tune the model using a consistent experimental setup for approximately 40 epochs. Our goal is to illustrate that models affected by contamination exhibit higher performance. Furthermore, evaluating benchmarks processed by our method \textit{Clean-Eval} aims to mitigate this performance inflation and restore the true capabilities of the LLMs.

\section*{Ethic Statement}
This paper will not pose any ethical problems. The datasets used in this paper have already been used in previous articles.

\section*{Acknowledgments}
Hongyuan is supported by FaceMind corporation, a grant from the Research Grant Council of the Hong Kong Special Administrative Region, China (Project Code: 14200719), and a grant from Center for Perceptual and Interactive Intelligence (CPII) Ltd. under the Innovation and Technology Commission’s InnoHK scheme. Rui is supported by the General Program of the National Natural Science Foundation of China (62176153), the Shanghai Municipal Science and Technology Major Project (2021SHZDZX0102), the Alibaba-AIR Program (22088682), and the Tencent AI Lab Fund RBFR2023012.

\bibliography{anthology,emnlp2023}
\bibliographystyle{acl_natbib}

\appendix

\section{Experiment Settings}
\label{sec:exp}
We conducted fine-tuning of the Llama2-7b-chat version on 2 RTX4090 GPUs, each with 24GB of memory. The model was fine-tuned according to specific instructions, utilizing the following prompt:

\begin{verbatim}
[INST] <<SYS>>\n"
"You are a helpful, respectful, and honest
assistant."
"<</SYS>>\n\n{0} [/INST]\n{1}</s>"]
\end{verbatim}

To optimize memory usage and enable deployment on smaller devices, we loaded our Llama2-7b-chat model in 4-bit precision, effectively reducing memory consumption. Employing a bfloat16 compute data type alongside nested quantization further contributed to memory efficiency. Additionally, we leveraged LoRA with a 16-dimensional updated matrix and scaling set at 64. A batch size 16 was chosen for shorter instructions, while longer instructions used a batch size of 4. The initial learning rate was set to 2e-4, coupled with the paged\_adamw\_8bit optimizer for training.

\section{Potential Contamination}
Table \ref{tab: date} shows the dates of the datasets we collected. In cases where the collection dates were not specified in the paper, we take the publication date. It is important to acknowledge that some new datasets may contain older data. In addition, the release date of the dataset may be earlier than the table.

\label{dataset_date}
\begin{table}[]
    \centering
    \renewcommand{\arraystretch}{1.3} 
    \begin{tabular}{p{0.4\linewidth}|p{0.4\linewidth}} \hline
    MRPC     &  March 3, 2005\\
    RTE     & April 11, 2005 \\
    WNLI &  March 21, 2011\\
    IMDB & June 19,  2011 \\
    SST2 & October 18, 2013 \\
    Agnews & September 9, 2015 \\
    SNLI & August 21, 2015 \\
    MultiArith & August 20, 2016\\
    QNLI & October 11, 2016 \\
    CNN-Dailymail & April 25, 2017 \\
    MNLI & February 19, 2018\\
    QQP & November 1, 2018\\
    BBC-XSUM & August 27, 2018\\
    COLA & October 1, 2019 \\
    PIQA & November 26, 2019 \\
    BOOLQ & May 24, 2019\\
    CB & July 25, 2019 \\
    MMLU & January 12, 2021 \\
    GSM8K & November 18, 2021\\
    CEVAL & November 6, 2023\\\hline 
    \end{tabular} 
    \caption{The date of each dataset}
    \label{tab: date}
\end{table}

Additionally, we gather information on the model release dates: text-davinci-003, launched in September 2021, and Llama2-7B, introduced on July 9, 2023.

In the text-davinic-003 report \cite{brown2020language}, they conducted data contamination experiments. Datasets include BOOLQ, PIQA, RTE, CB, and COPA. The dirty rates were 75.80, 89.90\%, 71.40\%, 100.0\%, and 100.0\%, respectively.

In the report for Llama2 \cite{touvron2023llama}, they conducted data contamination experiments. They pointed out that the degree of possible data contamination in the humanities and overall data in MMLU reached 94.5\% and 94.4\%, respectively. Therefore, we can assume that Llama2 included MMLU's data at the beginning of the training, which means that there may be data contamination.

\section{Prompt Design}

\subsection{Method prompt}
\label{sec:ed}
Our paraphrasing, back-translation, and equivalence detector prompts are shown in Table \ref{tab:model_prompt}.

\begin{table*}[!htp]
    \centering
    \renewcommand{\arraystretch}{1.3} % 增加行高
    \begin{tabular}{p{0.25\linewidth}|p{0.65\linewidth}} \hline
    \textbf{Method} & \textbf{Prompt Design}\\\hline
    \textbf{Paraphrase}     & Please paraphrase the following sentence without changing the meaning in 3 ways, then return as a list. \\
    \textbf{Back-translation}     & Please translate the following sentence into \textit{[language]} without changing the meaning. \\
    \textbf{Equivalence Detector} & Please determine whether the following sentences are equivalent.  \\\hline
    \end{tabular}
    \caption{Prompt designs of each method.}
    \label{tab:model_prompt}
\end{table*}

\subsection{Instruction for Each Dataset}
\label{data}
Our prompts for each benchmark are shown in Table \ref{tab:data_prompt}.

\begin{table*}[!htp]
    \centering
    \renewcommand{\arraystretch}{1.3} % 增加行高
    \begin{tabular}{p{0.25\linewidth}|p{0.65\linewidth}} \hline
    \textbf{Dataset} & \textbf{Prompt Design}\\\hline
    \textbf{RTE}     & The task is to determine whether a pair of sentences are entailed by each other. Just return entailment or not\_entailment.\\
    \textbf{QQP, MRPC}     & The task is to determine whether a pair of questions are semantically equivalent. Just return equivalent or not\_equivalent. \\
    \textbf{QNLI} & The task is to determine whether the context sentence contains the answer to the question. Just return entailment or not\_entailment. \\
    \textbf{MNLI, CB} & The task is to predict whether the premise entails the hypothesis, contradicts the hypothesis, or neither. Just return entailment, contradiction, or neutral. \\
    \textbf{WNLI} & The task is to predict if the sentence with the pronoun substituted is entailed by the original sentence. Just return entailment or not\_entailment. \\
    \textbf{SNLI} & The task is to determine whether a pair of sentences are entailed, contradicted, or neutral to each other. Just return entailment, contradiction, or neutral. \\
    \textbf{IMDB} & The task is to determine whether the sentiment of the text is positive or negative. Just return positive or negative. \\
    \textbf{PIQA} & The task is to select the best solution to the question. Just return the solution1 or solution2. \\
    \textbf{COPA} & Given a premise, choose one of the following two choices that express the {sample["question"]} relationship. Just return choice1 or choice2. \\
    \textbf{BOOLQ} &  The task is to answer true or false given the question. Just return true or false. \\
    \textbf{SST2} &  The task is to determine whether the sentiment of the sentence is positive or negative. Just return positive or negative.\\
    \textbf{AG News} &  The task is to classify the article into sports, world, business, or sci/tech. Just return sports, world, business, or sci/tech. \\
    \textbf{GSM8K, MultiArith} &  The task is to answer a given mathematical question. Just directly return the final number answer. \\
    \textbf{MMLU, CEVAL} &  Please select the best answer from the options according to the question. Just return one answer with A, B, C, or D. \\
    \textbf{CNN\_Dailymail, BBC\_XSUM} & Please summarize this article.\\
    \hline
    \end{tabular}
    \caption{Prompt designs of each benchmark.}
    \label{tab:data_prompt}
\end{table*}

\subsection{BLEURT Score}
\label{bs}
\begin{figure*}[!htp]
  \centering
  \begin{subfigure}{0.32\textwidth}
    \centering
    \includegraphics[width=\linewidth]{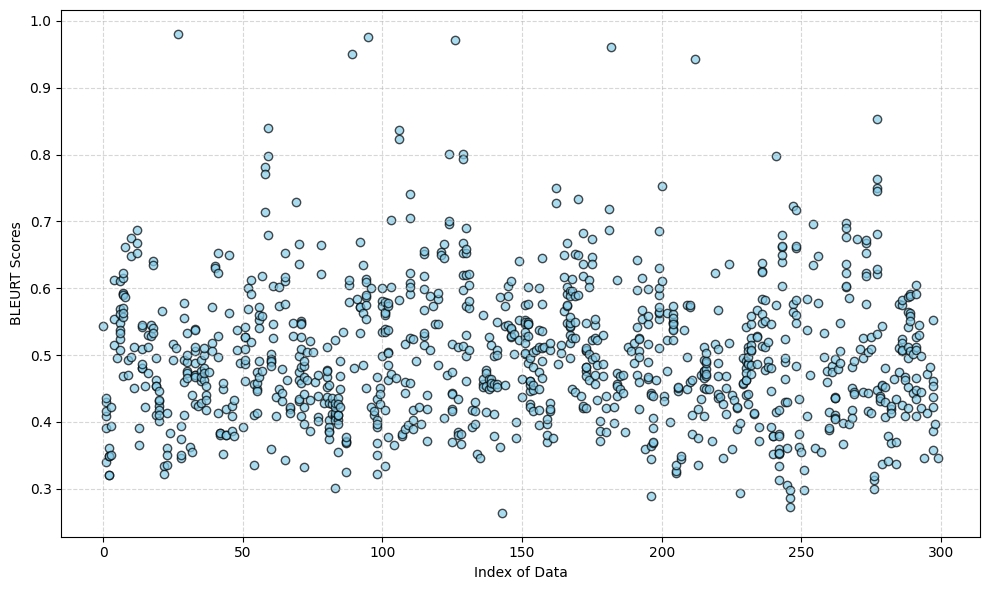}
    \caption{AgNews.}
    \label{fig:2a4}
  \end{subfigure}
  \hfill
  \begin{subfigure}{0.33\textwidth}
    \centering
    \includegraphics[width=\linewidth]{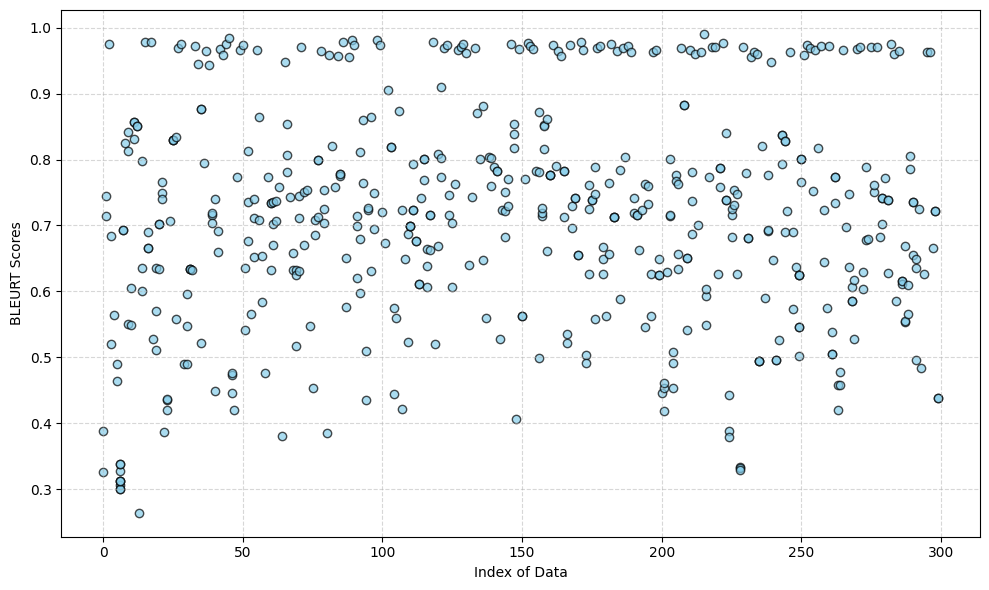}
    \caption{BOOLQ.}
    \label{fig:2b4}
  \end{subfigure}
  \begin{subfigure}{0.33\textwidth}
    \centering
    \includegraphics[width=\linewidth]{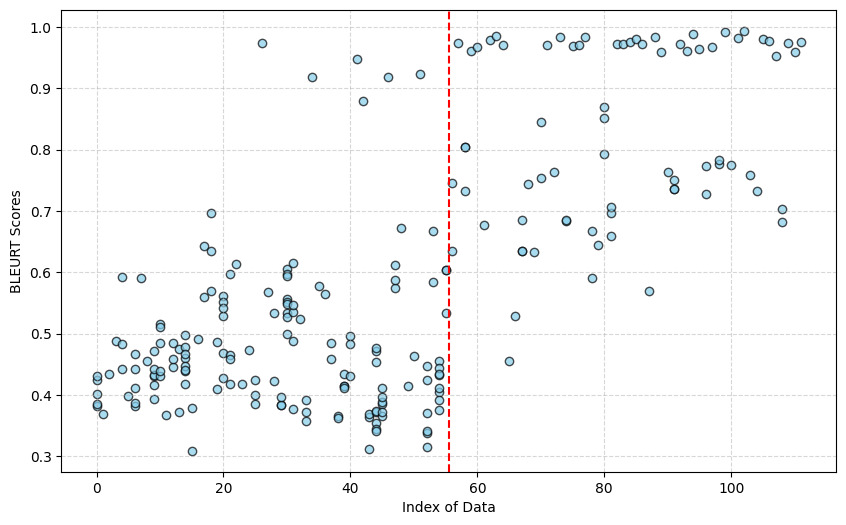}
    \caption{CB.}
    \label{fig:2c3}
  \end{subfigure}
  \centering
  \begin{subfigure}{0.32\textwidth}
    \centering
    \includegraphics[width=\linewidth]{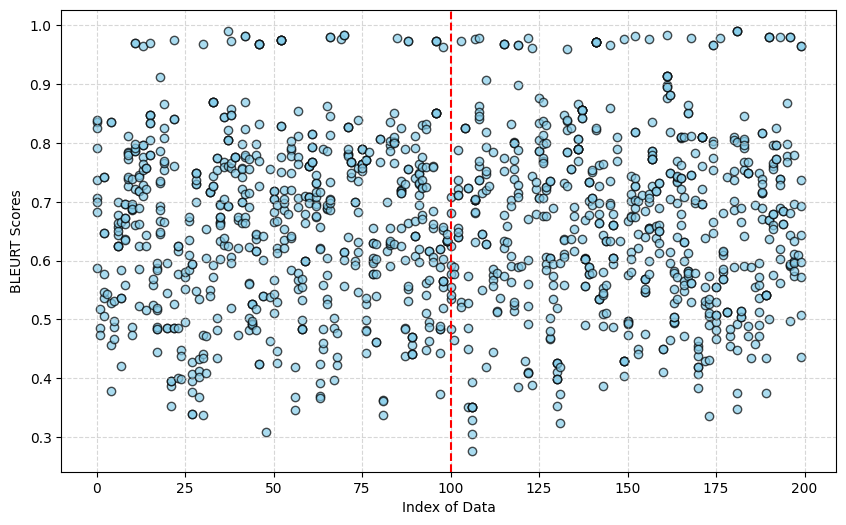}
    \caption{COPA.}
    \label{fig:2a3}
  \end{subfigure}
  \hfill
  \begin{subfigure}{0.33\textwidth}
    \centering
    \includegraphics[width=\linewidth]{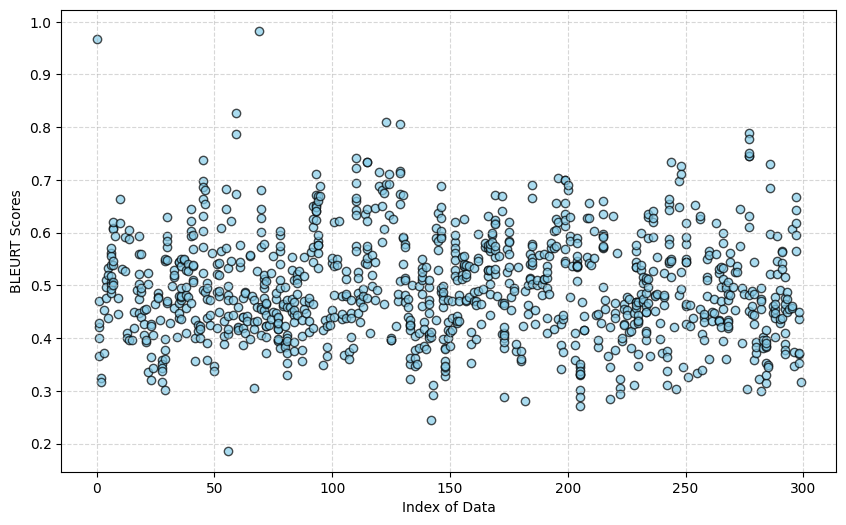}
    \caption{IMDB.}
    \label{fig:2b3}
  \end{subfigure}
  \begin{subfigure}{0.33\textwidth}
    \centering
    \includegraphics[width=\linewidth]{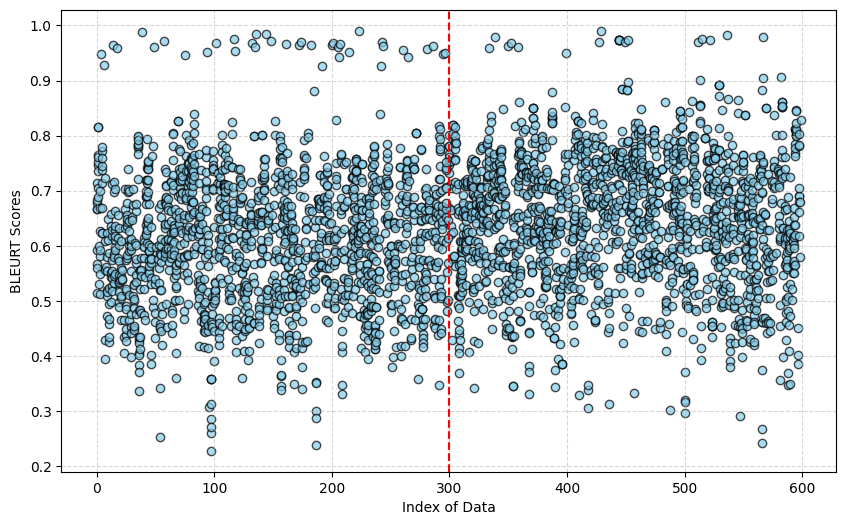}
    \caption{MNLI.}
    \label{fig:2c22}
  \end{subfigure}
  \centering
  \begin{subfigure}{0.32\textwidth}
    \centering
    \includegraphics[width=\linewidth]{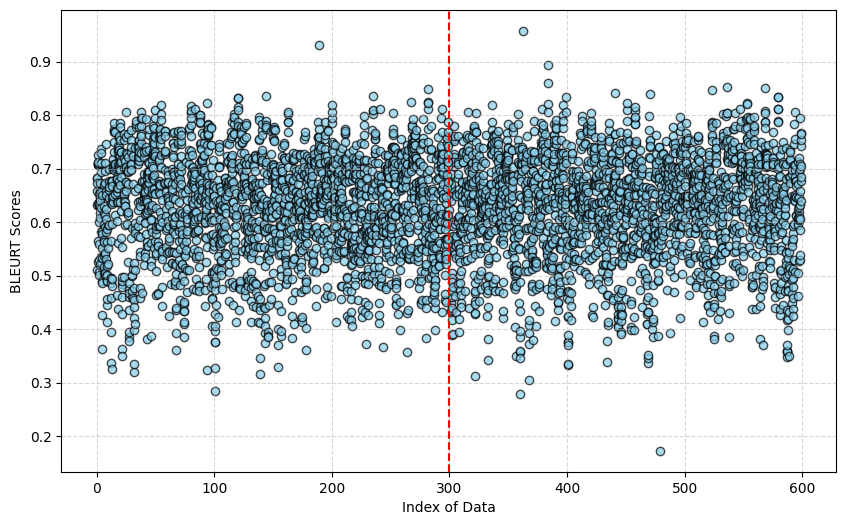}
    \caption{MRPC.}
    \label{fig:2a2}
  \end{subfigure}
  \hfill
  \begin{subfigure}{0.33\textwidth}
    \centering
    \includegraphics[width=\linewidth]{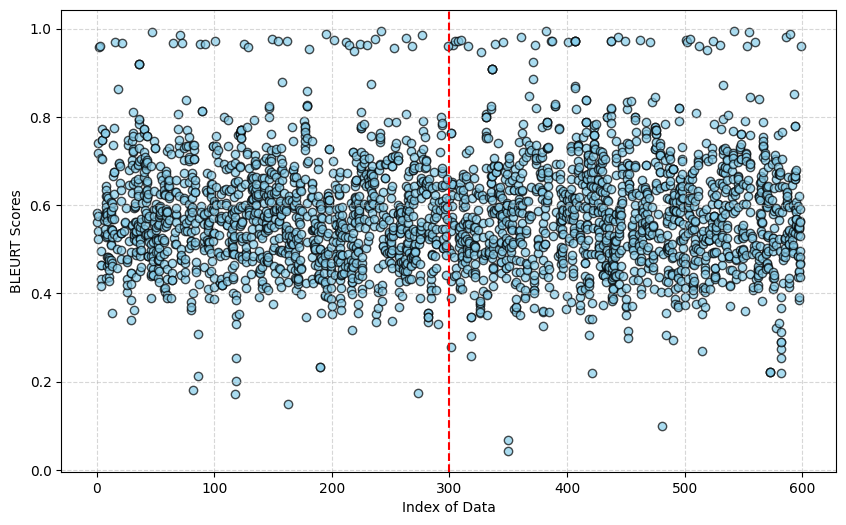}
    \caption{PIQA.}
    \label{fig:2b2}
  \end{subfigure}
  \begin{subfigure}{0.33\textwidth}
    \centering
    \includegraphics[width=\linewidth]{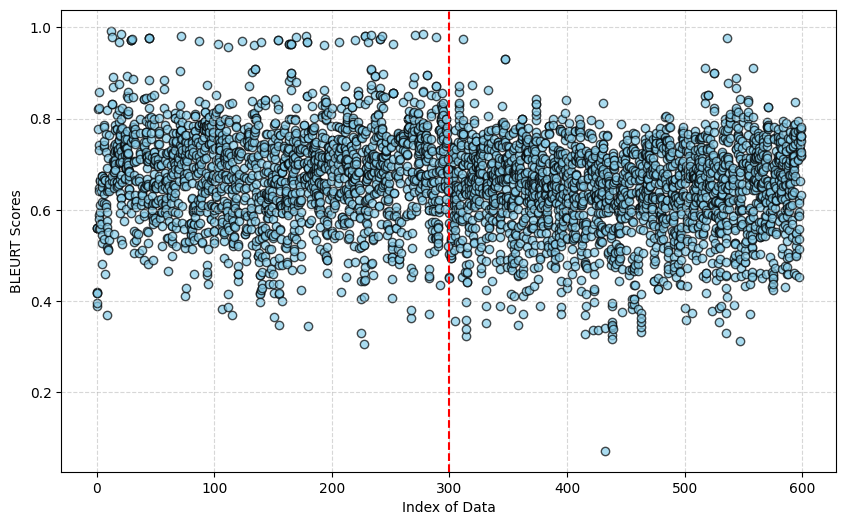}
    \caption{QNLI.}
    \label{fig:2c2}
  \end{subfigure}
  \centering
  \begin{subfigure}{0.32\textwidth}
    \centering
    \includegraphics[width=\linewidth]{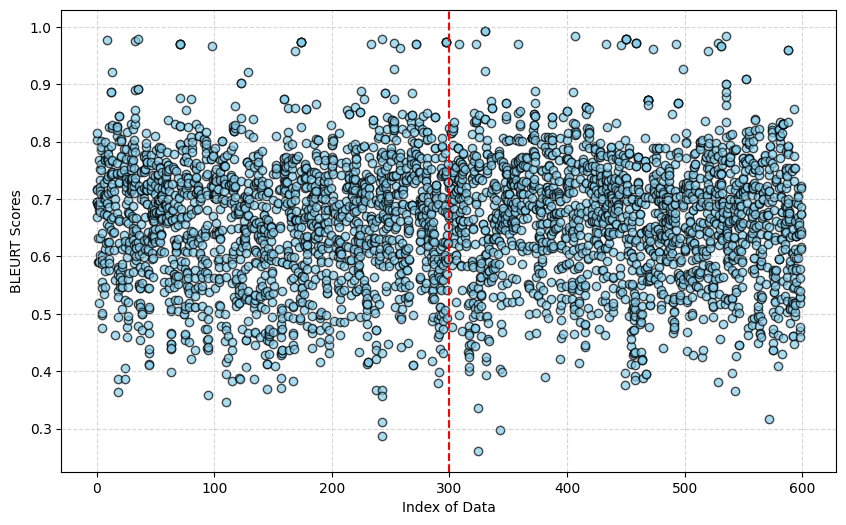}
    \caption{QQP.}
    \label{fig:2a1}
  \end{subfigure}
  \hfill
  \begin{subfigure}{0.33\textwidth}
    \centering
    \includegraphics[width=\linewidth]{img/PIQA.png}
    \caption{PIQA.}
    \label{fig:2b1}
  \end{subfigure}
  \begin{subfigure}{0.33\textwidth}
    \centering
    \includegraphics[width=\linewidth]{img/QNLI.png}
    \caption{QNLI.}
    \label{fig:2c1}
  \end{subfigure}
  \caption{The BLEURT score of each instance from selected benchmarks compared to the original. The graph featuring the red line represents a paired dataset, depicting one instance on either side of this demarcation}
  \label{fig:pipeline1}
\end{figure*}

Figure \ref{fig:pipeline1} illustrates the BLEURT score of each instance from selected benchmarks compared to the original instance.

\end{document}

%% file: abstract.tex
\begin{abstract}
We are currently in an era of fierce competition among various large language models (LLMs) continuously pushing the boundaries of benchmark performance. However, genuinely assessing the capabilities of these LLMs has become a challenging and critical issue due to potential data contamination. In this paper, we propose a novel and valuable method, \textit{Clean-Eval}, which mitigates the issue of data contamination and evaluates the LLMs more cleanly. \textit{Clean-Eval} employs a neural-based model to paraphrase and back-translate the contaminated data into a candidate set, generating expressions with the same meaning but in different surface forms. A semantic detector is then used to filter those generated low-quality samples to narrow down this candidate set. Candidates with moderate BLEURT scores against the original samples are selected as the final evaluation set. According to human assessment, this set is almost semantically equivalent to the original contamination set but expressed differently. We conduct experiments on 20 existing benchmarks across diverse tasks, and results demonstrate that \textit{Clean-Eval} substantially restores the actual evaluation results on contaminated LLMs under both few-shot learning and fine-tuning scenarios. %We will later be open-sourced as a website to measure LLMs fairly.
\end{abstract}

%% file: intro.tex
\section{Introduction}

In recent years, LLMs have made breakthroughs in handling complex and nuanced scenarios, achieved superior performance in some professional and academic benchmarks, and attracted many resources from industry and academia \cite{openai2023gpt4, touvron2023llama, golchin2023time}. This subsequently opens the arms race era of LLMs, and various LLMs are continuously launched, such as GPT-4 \cite{openai2023gpt4}, LLama2 \cite{touvron2023llama} and other LLMs, which have refreshed various evaluation benchmarks continuously.

\begin{figure}
    \centering
    \includegraphics[width=8cm]{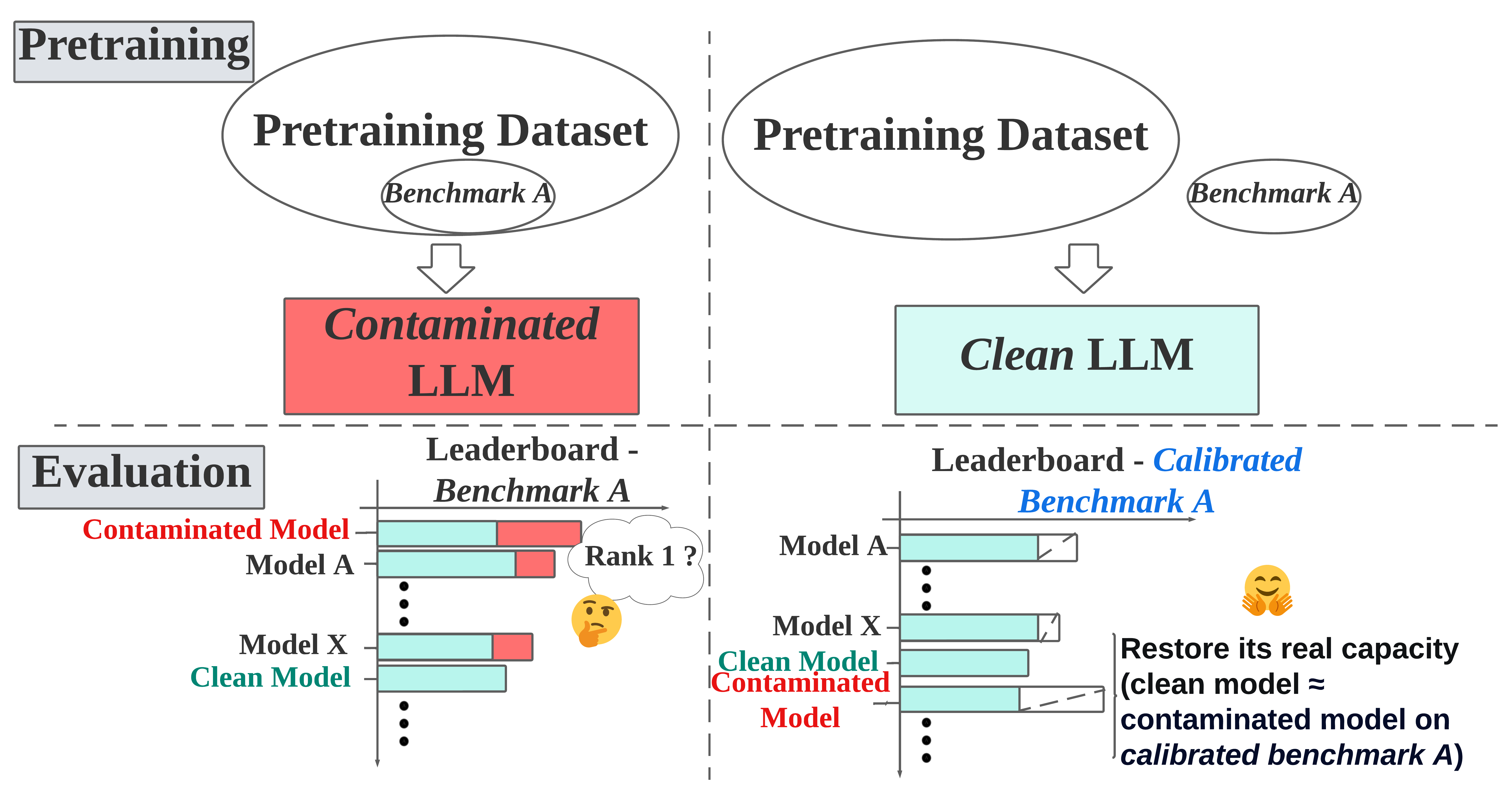}
    \caption{Data contamination happens when Benchmark A is included in the pretraining data, leading to inflated performance metrics like top leaderboard rankings.  This can cause a clean model to lag behind the contaminated one.  We aim to revise Benchmark A, preserving its meaning but changing its surface forms.  This aims to re-evaluate the contaminated model and align its performance closer to that of a clean model.}
    \label{fig:1}
\end{figure}

\begin{figure*}
  \centering
  \begin{subfigure}{0.32\textwidth}
    \centering
    \includegraphics[width=\linewidth]{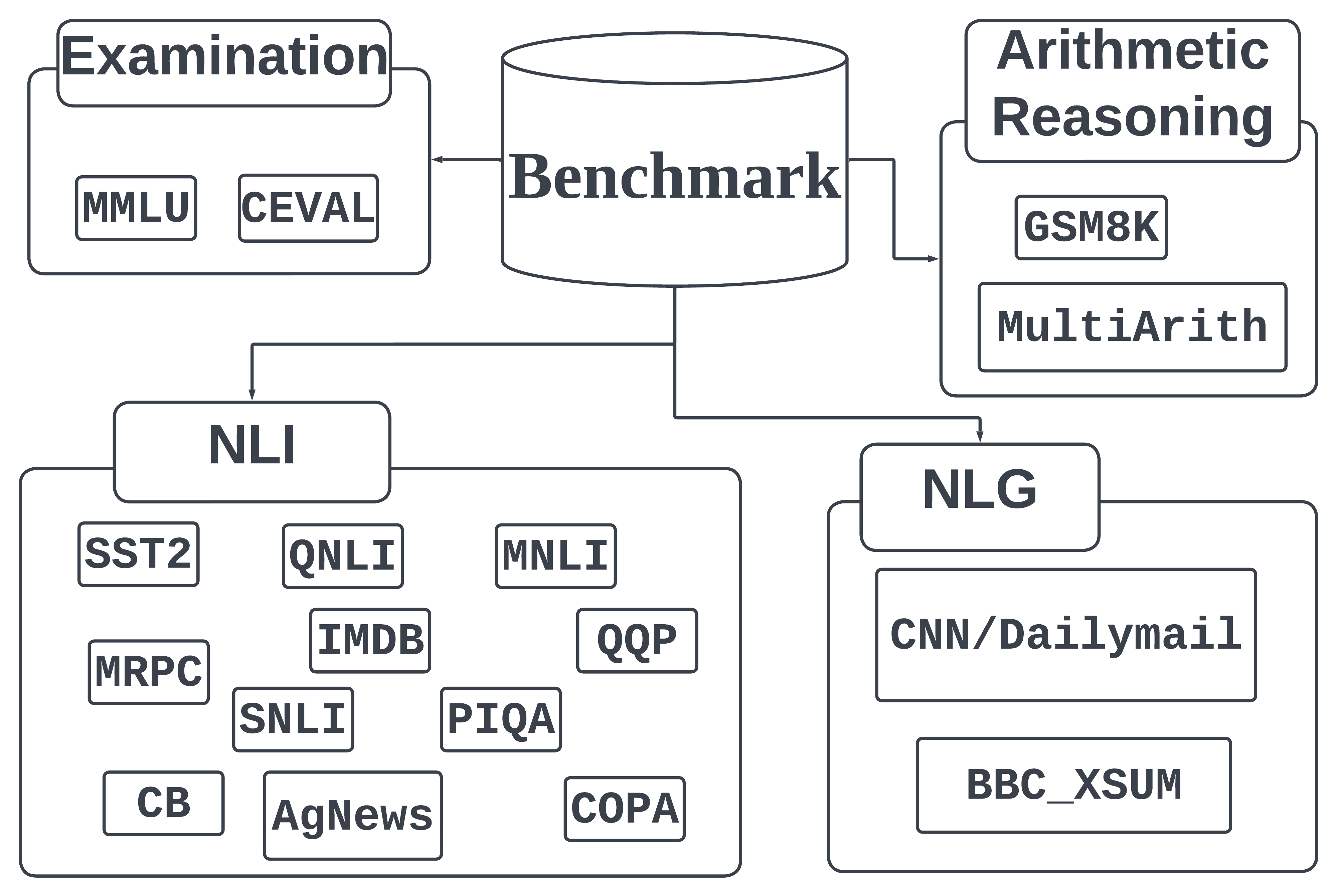}
    \caption{Data Collection.}
    \label{fig:2a}
  \end{subfigure}
  \hfill
  \begin{subfigure}{0.33\textwidth}
    \centering
    \includegraphics[width=\linewidth]{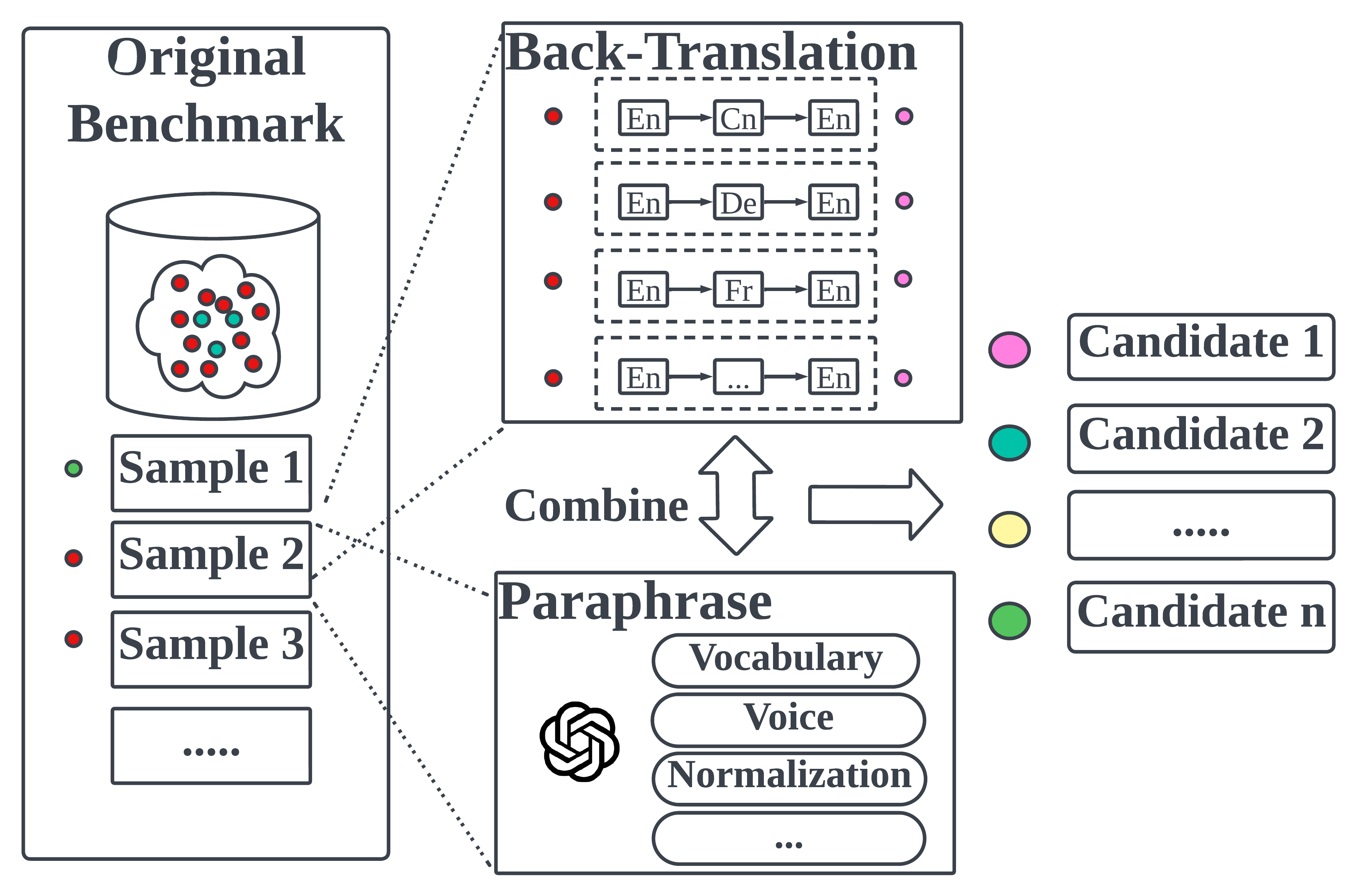}
    \caption{Paraphrase and Back-translation.}
    \label{fig:2b}
  \end{subfigure}
  \begin{subfigure}{0.33\textwidth}
    \centering
    \includegraphics[width=\linewidth]{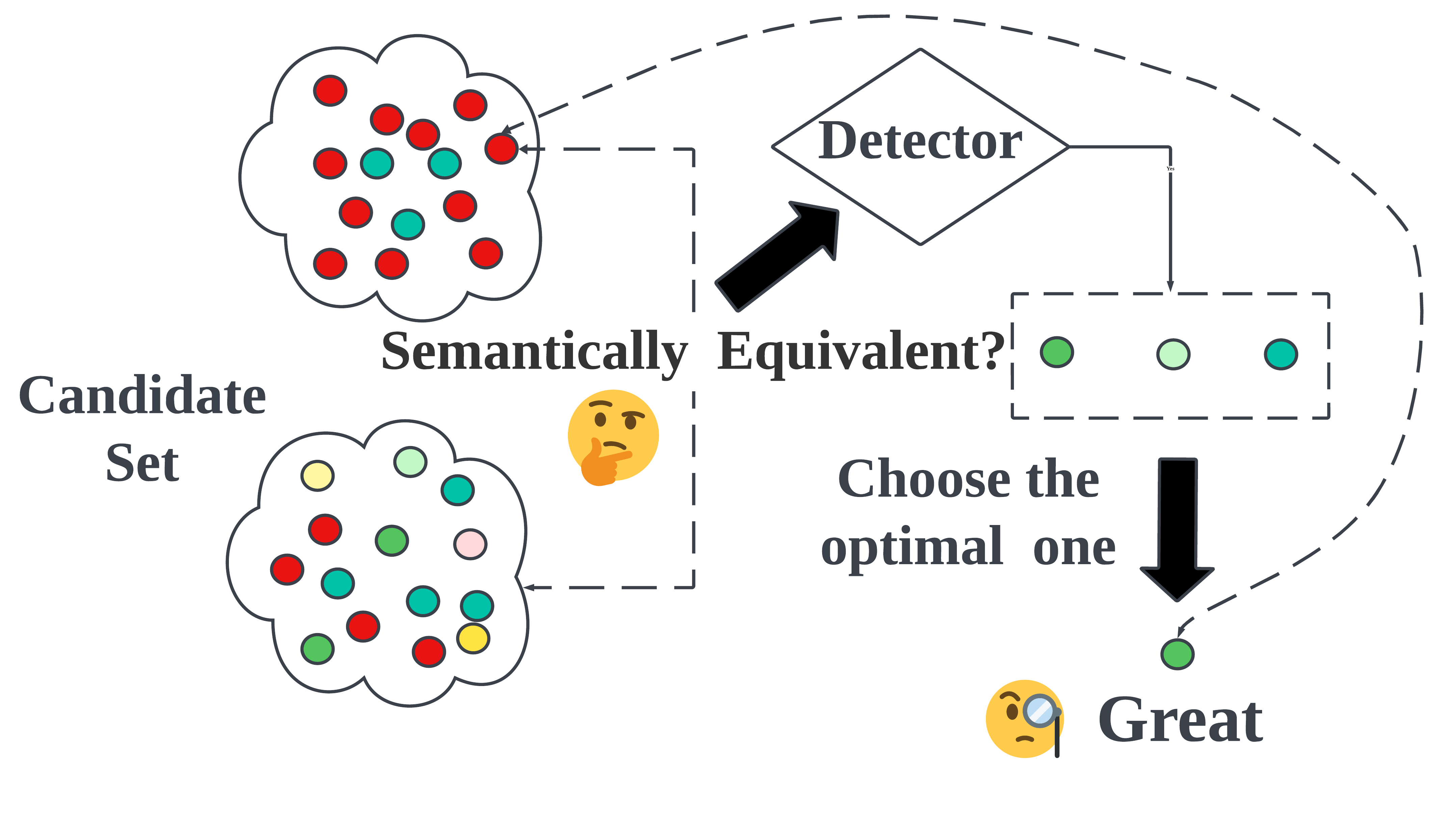}
    \caption{Semantic Filter.}
    \label{fig:2c}
  \end{subfigure}
  \caption{An overview of our method. We first gather existing benchmarks for LLM assessment and then meticulously clean contamination in these benchmarks through LLM-powered paraphrase and multi-language back-translation, employing a semantic detector to filter and select optimal results based on BLEURT scores.}
  \label{fig: pipeline}
\end{figure*}

There is room for doubt regarding the potential overestimation of these benchmark measurements. One reason is that LLMs are trained on data extracted from websites and publicly accessible datasets \cite{openai2023gpt4, touvron2023llama}. Therefore, ensuring no overlap between the pretraining dataset and the evaluated benchmark becomes challenging. This subsequently introduces a significant concern: the risk of data contamination.

Data contamination arises when the pre-training data of one model integrates evaluated data, consequently enhancing test performance \cite{magar2022data, golchin2023time}. Currently, many models opt not to disclose their training sets in technical reports, raising concerns about the potential inclusion of benchmark datasets within their training data. This presents an urgent problem \cite{wei2023skywork}, as these contaminated models claim highly evaluated results but often lead to poor real-world experiences. We strongly advocate for a cleaner evaluation of LLMs. Unveiling the genuine capabilities of LLMs could significantly propel the community of LLMs forward. The most effective resolution involves relabeling a new dataset when developing a new model to assess its capabilities. Unfortunately, this process demands considerable time and labor.

This paper employs previously proposed benchmarks to create a new benchmark, and our method is called \textit{Clean-Eval}, aiming to mitigate data contamination using LLMs and accurately assess the actual capabilities of LLMs. Leveraging the exceptional creative capabilities of these models, we perform diverse paraphrasing of contaminated data and back-translate it across multiple language directions. This process results in a pool of calibrated datasets. We effectively filter out low-quality samples by utilizing semantic detectors, and then select the best items based on BLEURT scores derived from comparisons between the calibrated and contaminated data. Finally, We conducted experiments on 20 benchmarks across diverse tasks, and our analysis unveiled noticeable calibrated effects achieved through \textit{Clean-Eval}. Our human evaluation reinforces the method's potential to improve sentence structure, grammar, and linguistic diversity while maintaining core semantics. Acknowledging the challenge of detecting model contamination within specific benchmarks, we propose a new evaluation approach for in-context learning and fine-tuning. Our experiments convincingly demonstrate that processing contaminated data through our method effectively restores the genuine performance of LLMs.

%% file: re.tex
\section{Related Work}

\subsection{Data Contamination}
Detecting data contamination is crucial in ensuring the integrity of model training and usage. Researchers and practitioners have dedicated considerable efforts to developing methods for identifying and mitigating instances where test data unintentionally becomes part of the training dataset of models \cite{brown2020language, touvron2023llama}.

\paragraph{Model Trainers.}
\citet{brown2020language} conducted experiments on data contamination, using an n-gram overlap metric to evaluate duplication levels between training and test sets. They subsequently eliminated these duplications from the training dataset. Similarly, \citet{dodge2021documenting} assessed exact matches, accounting for capitalization and punctuation normalization. This method scrutinized whether entire evaluation text inputs existed within the training data. However, \citet{touvron2023llama} critiqued the precision of previous high-order n-gram-based detection methods in determining contamination extent within a sample. Their proposed approach involved token-level contamination identification, allowing for slight variations in overlap positions between evaluation samples and training data. \citet{wei2023skywork} took a distinctive approach, comparing the LM loss between the test splits of a dataset and a mimic dataset generated by GPT-4 \cite{openai2023gpt4} to correspond to it. A smaller discrepancy value between these sets indicated potential contamination within the model.

\begin{figure*}
    \centering
    \includegraphics[width=15cm]{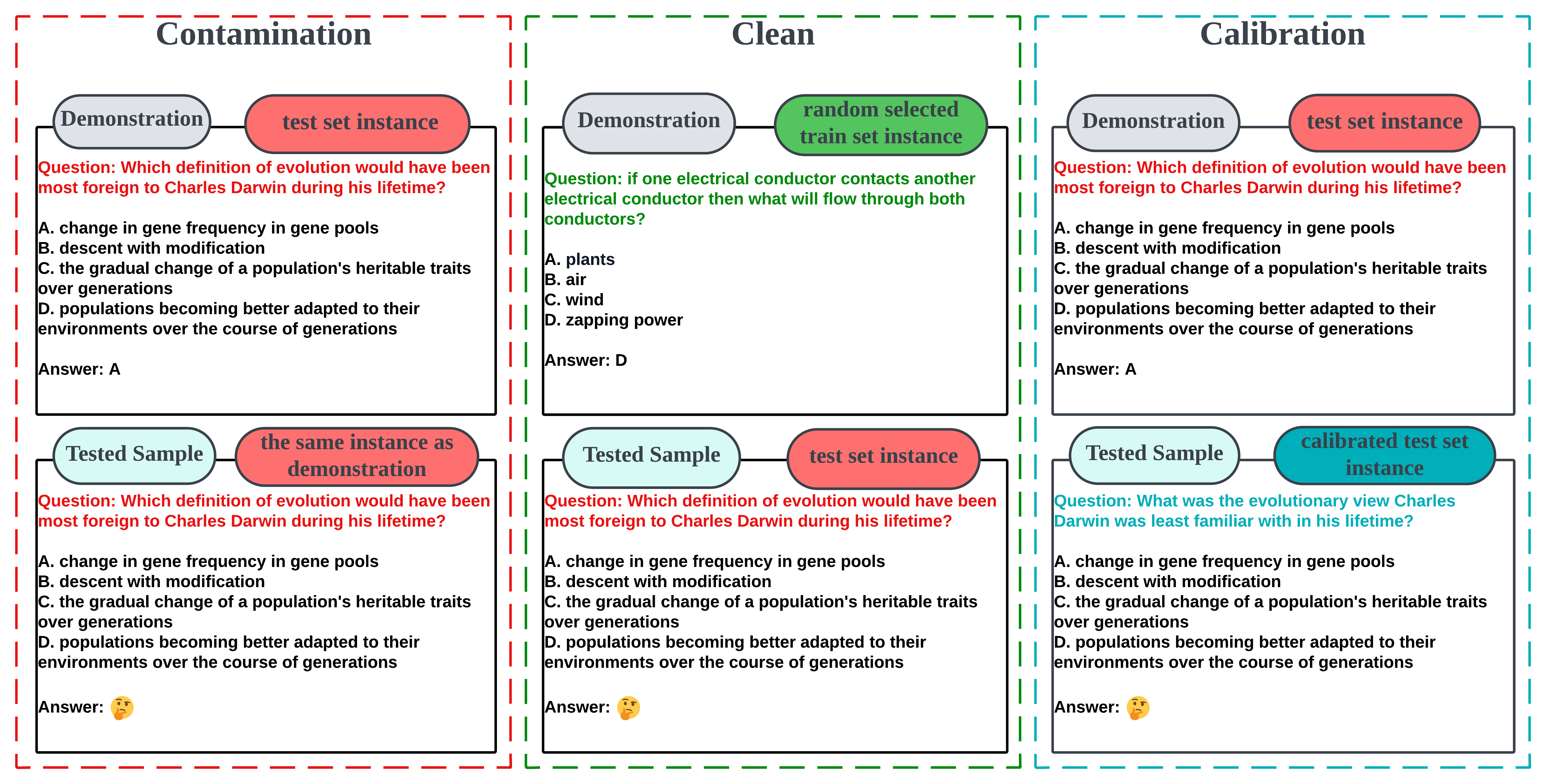}
    \caption{Evaluation setting of in-context learning. Each input comprises a demonstration and a tested sample. In the contamination setting, the demonstration matches the tested sample. In contrast, in the absence of contamination, the demonstration is drawn from another dataset split, maintaining distinction from the tested sample (e.g., sampled from the train split). In our Clean-Eval setup, the tested sample is a calibrated version of the demonstration, specifically designed to mitigate the effects of contamination.}
    \label{fig:es-ICL}
\end{figure*}

\paragraph{Model Users.} \citet{carlini2023quantifying} construct a set of prompts using the model's training data. They investigated by supplying prefixes of these prompts to the trained model to assess the model's capacity to complete the remaining portion of the example verbatim. Their study revealed that as the model's capacity, duplicated numbers, and context length increased, the models would be more proficient in memorizing data. Meanwhile, \citet{golchin2023time} introduced an approach involving the development of guided instructions that include the initial segment of a data instance and its corresponding partition name. These guided instructions are subsequently utilized to induce the model to generate the second part of the data, based on a provided prompt. Rouge \cite{lin-2004-rouge}, BLEURT \cite{sellam-etal-2020-bleurt}, and GPT4 auto evaluation determine whether the model had data contamination. Furthermore, \citet{li2023open} analyzed six prominent multi-choice QA benchmarks, quantifying their overlap with the training dataset already known of Llama to detect potential data contamination.

\subsection{Existing Benchmark}
Many benchmarks have been proposed, including MMLU \cite{li2023cmmlu}, CEVAL \cite{huang2023ceval}, etc., to measure the capability of LLMs comprehensively. However, labeling these benchmarks is time-consuming and laborious, and ensuring no overlap with the training set of LLM is often challenging.  There is also work to reformulate existing benchmarks to build new ones. For example, \citet{li2023reformeval} propose ReForm-Eval to reformulate existing benchmarks into unified large vision-language model compatible formats.

Nevertheless, based on our knowledge, there is no proposed solution to the problem of data contamination causing inflated model evaluation performance. In this paper, we propose an effective method to mitigate this problem. Experiments demonstrate that our methods work in evaluating both closed and open LLMs.

%% file: method.tex
\section{Clean-Eval}

\begin{table*}[!htp]
\footnotesize
\centering
\renewcommand{\arraystretch}{1.2}
\setlength{\tabcolsep}{3pt}
\begin{tabular}{l *{9}{Sc}}
\toprule
\rowcolor{gray!15}
\multicolumn{10}{c}{\textbf{text-davinci-003\quad|\quad In-context Learning \quad|\quad [Accuracy]}}\\
\midrule
& \textbf{AG News} & \textbf{QQP} & \textbf{QNLI} & \textbf{RTE} & \textbf{MNLI} & \textbf{WNLI} & \textbf{SNLI} & \textbf{IMDB} & \textbf{PIQA} \\
\midrule
w/ Contamination & 53.67 & 95.00 & 90.67 & 96.39 & 80.00 & 95.78 & 94.00 & 95.67 & 86.33\\
Possibly w/o Contamination & 40.67 & 83.33 & 80.00 & 84.12 & 71.00 & 54.93 & 73.67 & 89.00 & 80.33 \\
Clean-Eval & 53.00 \textcolor{red}{$\downarrow$} & 79.00 \textcolor{red}{$\downarrow$}  & 82.00 \textcolor{red}{$\downarrow$}& 76.90 \textcolor{red}{$\downarrow$}& 71.67\textcolor{red}{$\downarrow$} & 71.83 \textcolor{red}{$\downarrow$}&  62.00 \textcolor{red}{$\downarrow$}& 85.33 \textcolor{red}{$\downarrow$} & 75.33 \textcolor{red}{$\downarrow$}\\
\midrule
& \textbf{MultiArith} & \textbf{MRPC} & \textbf{GSM8K} & \textbf{COPA} & \textbf{CB} & \textbf{BOOLQ} 
 & \textbf{SST2} & \textbf{MMLU} & \textbf{CEVAL}\\
\midrule
w/ Contamination & 65.00 &  93.67 & 64.33 & 92.00 & 98.21 & 87.33 & 90.67 & 73.67 & 66.33\\
Possibly w/o Contamination & 35.00 & 68.33 & 12.33 & 90.00 & 82.14 & 81.33 & 80.00 & 59.00 & 41.00   \\
Clean-Eval & 60.00\textcolor{red}{$\downarrow$} & 65.67\textcolor{red}{$\downarrow$} & 50.67 \textcolor{red}{$\downarrow$} & 75.00\textcolor{red}{$\downarrow$}  & 91.07 \textcolor{red}{$\downarrow$} & 83.67 \textcolor{red}{$\downarrow$}  & 78.00\textcolor{red}{$\downarrow$} & 57.00 \textcolor{red}{$\downarrow$} & 38.33 \textcolor{red}{$\downarrow$} \\
\rowcolor{gray!15}

\toprule
\rowcolor{gray!15}
\multicolumn{10}{c}{\textbf{Llama2\quad|\quad \quad Fine-Tuning \quad\quad|\quad [Accuracy]}}\\
\midrule
& \textbf{AG News} & \textbf{QQP} & \textbf{QNLI} & \textbf{RTE} & \textbf{MNLI} & \textbf{WNLI} & \textbf{SNLI} & \textbf{IMDB} & \textbf{PIQA} \\
\midrule
w/ Contamination & 54.00 &  99.00 & 98.00 & 99.27 & 99.67 & 63.38 & 99.00 & 97.33 & 100.00\\
Possibly w/o Contamination & 31.67 & 84.00 & 85.67  & 80.51 & 72.00 & 47.89 & 82.00 & 94.00 & 74.33\\
Clean-Eval & 51.34 \textcolor{red}{$\downarrow$}  & 81.00 \textcolor{red}{$\downarrow$} & 79.00 \textcolor{red}{$\downarrow$} & 67.87\textcolor{red}{$\downarrow$} & 73.67 \textcolor{red}{$\downarrow$}  & 60.56 \textcolor{red}{$\downarrow$} & 68.37 \textcolor{red}{$\downarrow$} & 95.33 \textcolor{red}{$\downarrow$}& 78.67\textcolor{red}{$\downarrow$}  \\
\midrule
& \textbf{MultiArith} & \textbf{MRPC} & \textbf{GSM8K} & \textbf{COPA} & \textbf{CB} & \textbf{BOOLQ} 
 & \textbf{SST2} & \textbf{MMLU} & \textbf{CEVAL}\\
\midrule
w/ Contamination& 36.11 & 96.33 & 50.67 & 100.00 & 85.71 & 99.33  & 99.99 & 82.67 & 87.33 \\
Possibly w/o Contamination & 16.11 & 79.33 & 7.00 & 89.00 & 58.93 & 73.33 & 94.67 & 37.33 & 30.00\\
Clean-Eval  & 22.78 \textcolor{red}{$\downarrow$}& 60.33 \textcolor{red}{$\downarrow$} & 26.33\textcolor{red}{$\downarrow$}  & 76.00 \textcolor{red}{$\downarrow$}& 71.43 \textcolor{red}{$\downarrow$} & 91.33\textcolor{red}{$\downarrow$} & 90.67 \textcolor{red}{$\downarrow$} & 25.00  \textcolor{red}{$\downarrow$}  & 85.00 \textcolor{red}{$\downarrow$}  \\
\bottomrule
\end{tabular}
\caption{Natural language understanding tasks. The symbol \textcolor{red}{$\downarrow$} indicates a decrease in performance compared to the contamination setting. The optimal candidate is chosen according to the lowest BLEURT score.}
\label{tab:1}
\end{table*}

The framework of our method is shown in Figure \ref{fig: pipeline}. Our methodology comprises three primary stages. Initially, we concentrate on gathering existing benchmarks to assess LLMs. In the subsequent phase, we meticulously cleaned contamination in the collected benchmarks. This involves paraphrasing samples using the creative capacities of the LLMs and performing multi-language back-translation on the contaminated data. In the final phase, we use the semantic detector to filter the outcomes of the contamination cleanup, eliminating subpar results and selecting the ultimate results based on the BLEURT score. 

\subsection{Back-translation}

Back-translation (BT) involves retranslating content from the target language into its source language using literal terms \cite{sennrich-etal-2016-improving}. In this process, slight differences can be introduced, such as replacing synonyms. Therefore, we translate the raw data into various language orientations and then revert to the original language to compose our candidate set of contamination cleanup data. In this process, we aim to achieve a distinct expression from the original sample while preserving the semantics.

\subsection{Paraphrase}
LLMs have showcased significant potential across diverse professional domains, particularly in creative writing \cite{touvron2023llama}. Harnessing their creative prowess, we utilize LLMs to generate multiple paraphrases of raw data, purposefully introducing variations. Specifically, we leverage the text-davinci-003 version of GPT-3 to generate these paraphrases. For instance, a typical prompt in our approach was: \textit{Please paraphrase this sentence in three different ways.}

\subsection{Filter}

However, these candidate sets might need further examination to ensure their quality. As shown in Figure \ref{fig:2c}, we use a semantic detector to judge whether the content in the candidate set is semantically similar to the original content to narrow the set of candidate sets further and select the candidate according to the BLEURT score as the final result.\footnote{This detector is optional. Removing the detector saves computational and token costs, but can potentially degrade the quality of the selected candidates.} In Appendix \ref{bs}, the BLEURT scores of each instance on various benchmarks are presented, with scores typically ranging from 0.4 to 0.9. Our analysis indicates that the lowest BLEURT score is an effective indicator for restoring the true capabilities of LLMs.  
 
With these essential steps, we have achieved greater efficiency in harnessing existing datasets, mitigated data contamination concerns, and furnished high-calibrated new data suitable for evaluating model performance.

%%%%%%%%%%%%%%%%%%%%%%%%%%%%%%%%%%%%%%%%%%%%%%%%%%%%%%%%%%%%%%%%%%%%%%%%

\section{Evaluation Setting}

\begin{table*}[!htp]
\footnotesize
\centering
\renewcommand{\arraystretch}{1.2}
\setlength{\tabcolsep}{5pt} 
\begin{tabular}{l *{9}{Sc}}
\toprule
% \multicolumn{10}{c}{\textbf{text-davinci-003\quad|\quad In Context Learning \quad|\quad [Rouge]}}\\
% \midrule
& & \multicolumn{3}{c}{\textbf{CNN/Daily-Mail}} &  & \multicolumn{3}{c}{\textbf{BBC-XSUM}} \\
\cmidrule(lr){2-5} \cmidrule(lr){6-9}
& & \textbf{Rouge-1} & \textbf{Rouge-2} & \textbf{Rouge-L} &  & \textbf{Rouge-1} & \textbf{Rouge-2} & \textbf{Rouge-L}\\
\midrule
w/ Contamination & & 23.38 & 9.45 & 21.69 &  & 33.64 & 18.56 & 29.2\\
Possibly w/o Contamination & & 21.18 & 7.18 & 19.57 &   & 22.97 & 7.78 & 19.08 \\
Clean-Eval &  & 23.14 \textcolor{red}{$\downarrow$} & 9.35 \textcolor{red}{$\downarrow$} & 21.41 \textcolor{red}{$\downarrow$} &  & 33.09\textcolor{red}{$\downarrow$} & 17.92\textcolor{red}{$\downarrow$} & 28.90 \textcolor{red}{$\downarrow$}\\
\bottomrule
\end{tabular}
\caption{ICL experiments and metrics in Rouge. \textcolor{red}{$\downarrow$} is compared to the contamination dataset. The optimal candidate is chosen according to the lowest BLEURT score.}
\label{tab:2}
\end{table*}

\begin{figure}
    \centering
    \includegraphics[width=7.5cm]{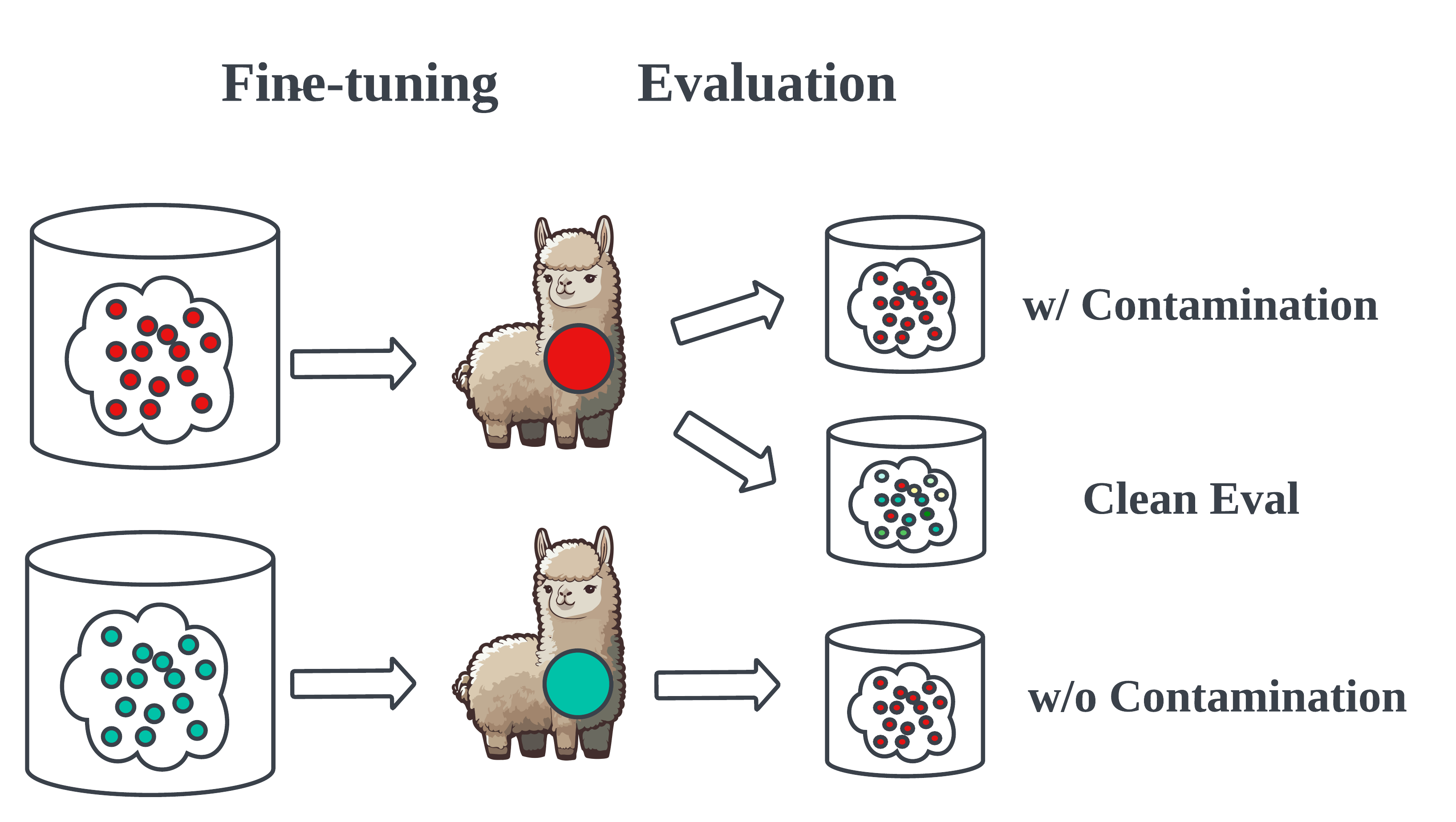}
    \caption{Evaluation setting of fine-tuning. We fine-tuned two models using datasets labeled red and green. When evaluated on the red dataset, these two models are categorized as contaminated and uncontaminated. Testing a model's performance on the red dataset processed by \textit{Clean-Eval} is attributed to the Clean-Eval setting. }
    \label{fig:es-ft}
\end{figure}

Nearly all LLMs operate with proprietary training datasets, making it challenging to ascertain whether the data being tested is free from contamination. We introduce an experimental framework for simulating data contamination to address this issue.

\subsection{In-context Learning}

\label{section4.1:ICL}

In-context learning (ICL) involves presenting a task demonstration to the model as a part of a natural language prompt. According to \citet{brown2020language}, LLMs are classified as few-shot learners. Due to restricted access to the GPT-3 model and its variability, we execute ICL on these models to assess the efficacy of \textit{Clean-Eval}. Within the ICL scenario, we propose and compare three evaluation settings: contamination, possibly no contamination, and clean evaluation for any given benchmark.

Each input comprises a demonstration and a tested sample, with different evaluation settings contingent upon their constitution. The demonstration matching the tested sample, depicted on the left side of Figure \ref{fig:es-ICL}, constitutes the \textbf{contamination setting}. When the demonstration and tested sample originate from different dataset splits (center of Figure \ref{fig:es-ICL}), it is categorized as the \textbf{possibly without contamination setting}. In contrast, when the tested sample is the demonstration processed by \textit{Clean-Eval} (right side of Figure \ref{fig:es-ICL}), it represents the \textbf{Clean-Eval setting}.

\subsection{Fine-tuning}

Fine-tuning entails further optimization adjustments for a specific task or dataset using a pre-trained LLM. Illustrated in Figure \ref{fig:es-ft}, we fine-tune two models using distinct splits of a dataset.

Each instance within a benchmark is formatted as an instruction for fine-tuning the model.
When the evaluation data mirrors the fine-tuned data, it's categorized as the \textbf{contamination setting}. If the evaluation and fine-tuned data originate from different splits of the same dataset, it falls under the \textbf{possibly without contamination setting}. Lastly, when the evaluation data is fine-tuned data processed by \textit{Clean-Eval}, it represents our \textbf{Clean-Eval setting}.

%% file: exp.tex
\section{Experiments}

\begin{table*}[!htp]
    \centering
    \footnotesize
    \renewcommand{\arraystretch}{1.1}
    \setlength{\tabcolsep}{6pt}
    \begin{tabular}{lcccccc}\toprule
    \textbf{Data} & \textbf{Method} & \textbf{Rouge-1} & \textbf{Rouge-2} & \textbf{Rouge-L} & \textbf{BLEURT} & \textbf{Equivalence} \\\midrule
    \multirow{3}{*}{QNLI} & Back-translation & 54.08 & 29.80 & 50.47 & 63.44 & 100.00\\
    & Paraphrase & 48.50 & 26.02 & 43.28 & 63.19 & 100.00\\
    & Clean-Eval & \textbf{46.85} & \textbf{22.90} & \textbf{42.53} & \textbf{60.21} & 100.00 \\\midrule
    \multirow{3}{*}{SST2} & Back-translation & 52.35 & 32.05 & 51.01 & 59.94 & 100.00\\
    & Paraphrase & 30.39 & 9.98 & 27.77 & 42.96 & 100.00 \\
    & Clean-Eval & \textbf{26.66} & \textbf{7.55} & \textbf{23.64} & \textbf{40.90} & 100.00\\\midrule
    \multirow{3}{*}{MMLU} & Back-translation & 52.85 & 30.15 & 48.68 & 57.91 & 100.00 \\
    & Paraphrase & 45.71 & 23.10 & 40.79 & 55.46 & 100.00 \\
    & Clean-Eval & \textbf{42.42} & \textbf{19.70} & \textbf{38.32} & \textbf{51.99} & 100.00\\
    \bottomrule  
    \end{tabular}
    \caption{The difference between the sample processed with different methods and the original sample. We choose the lowest BLEURT score as our optimal candidate. As all generated samples undergo semantic detection, their semantic equivalence consistently reaches 100\%.}
    \label{tab:automatic-evaluation}
\end{table*}

\subsection{Datasets}
We have meticulously curated 20 datasets, spanning a wide array of tasks. These tasks encompass text implication, problem pair matching, natural language reasoning, semantic similarity, sentiment analysis, common sense reasoning, text classification, mathematical reasoning, examinations, and even some natural language generation tasks. This classification provides valuable insights into the performance of various task types concerning data contamination. Below is the comprehensive list of datasets we have utilized. The specific release date of the dataset is listed in the Appendix \ref{dataset_date}.

\begin{itemize}
    \item \textbf{Nature Language Inference.} GLUE dataset \cite{wang2019glue} that includes QNLI, MNLI, SNLI, WNLI, RTE, QQP, MRPC, SST2; IMDB \cite{maas-EtAl:2011:ACL-HLT2011}; BOOLQ \cite{clark2019boolq}; Super-GLUE dataset \cite{wang2019superglue} that includes COPA, CB; Ag News \cite{Zhang2015CharacterlevelCN}.
    \item \textbf{Nature Language Generation.} CNN\_Dailymail \cite{see-etal-2017-get}, BBC\_XSUM \cite{Narayan2018DontGM}.
    \item \textbf{Arithmetic Reasoning.} GSM8K \cite{cobbe2021gsm8k}, MultiArith
    \item \textbf{Examination.} MMLU \cite{hendryckstest2021}, CEVAL \cite{huang2023ceval}.
\end{itemize}

\subsection{Metrics}
\paragraph{ROUGE \& BLEURT.} To measure the degree of overlap between a generated instance and a reference, we utilize both ROUGE \cite{lin-2004-rouge}, and BLEURT scores \cite{sellam-etal-2020-bleurt}. ROUGE evaluates lexical similarity, focusing on shared words and phrases, while BLEURT assesses the semantic relevance and fluency of the generated sequence concerning the reference instance. 

\paragraph{Equivalence.} We employed the text-davinci-003 model \cite{brown2020language} to assess equivalence before and after the processing of contaminated data by \textit{Clean-Eval}. Details of the prompt designs are in Appendix \ref{sec:ed}.

\subsection{Contamination Cleanup.} 

\paragraph{Models.} We employ the text-davinci-003 model \cite{brown2020language} for paraphrasing, back-translation, and semantic detection purposes. We also utilize the BLEURT-20 model \cite{sellam-etal-2020-bleurt} to compute BLEURT scores and select the optimal candidate.

\paragraph{Process.}  Given the diversity in format and content across datasets, our processing criteria vary accordingly. Resource constraints prevent comprehensive processing of every dataset aspect within our method, \textit{Clean-eval}. For instance, while we thoroughly handle all contents in SNLI-paired datasets, our focus narrows to questions alone in question-options-answer or question-answer datasets. Additionally, our analysis is limited to the initial three sentences or less when dealing with lengthy text. Furthermore, all generated samples undergo semantic detection. If they fail this detection, the original sample is output.

\paragraph{Results.} The results are shown in Table \ref{tab:automatic-evaluation}. Following our \textit{Clean-Eval} method, the surface form of the newly generated sample notably differs from the original sample, particularly in terms of n-gram variations. However, the presence of the semantic detector ensures the quality and fidelity of the generated results, assuring their reliability despite these surface-level alterations.

\subsection{In-context Learning}

\paragraph{Model.} 
We use the text-davinci-003 model \cite{brown2020language} to conduct ICL experiments. 

\paragraph{Implementation Details.}
Each tested use case is provided with task-specific instructions. For instance, one instance attributed to CNN/Dailymail would receive a prompt such as ``The task is to summarize this article:''. Detailed designs for all prompts are in Appendix \ref{data}.

\paragraph{Results and Analysis}
The results displayed in Table \ref{tab:1} and Table \ref{tab:2} consistently showcase superior performance across all tasks in the presence of data contamination, surpassing the possible no-contamination and Clean-Eval settings. This emphasizes a distinct performance advantage influenced by data contamination. Notably, the model demonstrates robust generalization across more straightforward tasks like RTE, IMDB, and QQP, which is evident from its strong performance even without possible contamination. However, when contamination occurs in these tasks, the model sustains a near-optimal performance level.

The Clean-Eval setting is reliable, revealing the model's genuine capability. Many datasets exhibit performance levels close to those without contamination. Yet, a performance gap exists between the possible no-contamination and Clean-Eval settings, especially in more intricate tasks involving mathematical reasoning, such as GSM8K and MultiArith. The model's reduced performance in the possible no-contamination setting might stem from a lack of chain of thought, leading to performance degradation. Moreover, as depicted in Table \ref{tab:2}, our approach effectively mitigates data contamination, even when limiting processing to the first three sentences or fewer in an article. All results indicate that employing our \textit{Clean-Eval} method results in a gradual performance decline, aligning more closely with the possible no-contamination setting.

\subsection{Fine-tuning}
\paragraph{Model.}
For fine-tuning, we employ the LLama2-7b-chat model \cite{touvron2023llama}.

\paragraph{Implementation Details}

As model parameters grow in size, achieving full fine-tuning becomes increasingly challenging. In such scenarios, we resort to LoRA for fine-tuning \cite{hu2021lora}. Additional experiment settings are detailed in Appendix \ref{sec:exp}. Our process commences by transforming original data into instructional data, followed by single-instruction fine-tuning. Considering the extensive datasets, conducting exhaustive fine-tuning for each model to attain optimal performance would be impractical and time-consuming. Thus, we fine-tune the model for approximately 40 epochs before assessing its performance.

\paragraph{Results and Analysis}

The results are displayed in Table \ref{tab:1}. When the model undergoes fine-tuning and subsequent performance testing using the same dataset, it achieves notably higher accuracy, even reaching $100\%$ on some datasets. However, this performance dips when evaluated on a different dataset split. A significant performance gap exists between the possibly uncontaminated and contaminated dataset settings, particularly in challenging tasks like MultiArith, GSM8k, MMLU, and CEVAL. Notably, when tested under a Clean-Eval setting, the model's performance aligns closely with the possibly uncontaminated data.

%% file: ana.tex
\section{Analysis}

\subsection{Ablation Study}
In Table \ref{tab:automatic-evaluation}, we conducted an ablation study comparing three methods, including back-translation, paraphrase, and \textit{Clean-Eval}. Back-translation consistently yields higher Rouge and BLEURT scores than other methods across three datasets. This suggests that back-translation effectively maintains lexical and sentence structure from the original text. Paraphrase introduces variations in content expression, showcasing the ability to offer alternative ways of expressing the same semantic content. \textit{Clean-Eval}, which combines paraphrase and back-translation, emerges as a comprehensive approach. It maintains semantic equivalence, as indicated by the Equivalence score, and enhances the diversity of content expression.

\subsubsection{BLEURT Score}

In this part, we explored whether the selection based on the BLEURT score impacts the model performance.
 
\begin{table}[!htp]
    \centering
    \footnotesize
    \renewcommand{\arraystretch}{1.1} 
    \setlength{\tabcolsep}{4pt} 
    \begin{tabular}{lcccc}\toprule
    \textbf{Method} & \textbf{Score} & \textbf{QNLI} & \textbf{SST2} & \textbf{MMLU} \\ \midrule
    \multirow{3}{*}{BT} & lowest & \textbf{7.33} & \textbf{6.00} & \textbf{14.67}\\
    & median & -5.99 & 6.00 & 3.33\\
    & highest & -10.67 & 6.00 & 4.01\\\midrule
    \multirow{3}{*}{Para} & lowest & -8.67 & \textbf{6.00} & 8.01 \\
    & median & \textbf{4.01} & 4.66 & 5.33 \\
    & highest & 0.01 & 6.00 & \textbf{8.67}\\
    \bottomrule\end{tabular}
    \caption{In ICL experiments, we assess the performance gap using various BLEURT scores.  This gap represents the difference in performance between the model tested in the Clean-Eval setting versus the no-contamination setting and the model tested in the contamination setting versus the Clean-Eval setting.   A higher value signifies that Clean-Eval approaches performance levels like those in the no-contamination setting.
}
    \label{tab:bleurt_score}
\end{table}

\paragraph{Results.}  Table \ref{tab:bleurt_score} illustrates that paraphrasing exhibits variability across three datasets. However, back-translation demonstrates the potential to bring the model's performance closer to the no-contamination setting when choosing the lowest BLEURT score. Hence, to restore the large model's capabilities, selecting the best candidate based on the lowest BLEURT score might be a viable strategy.

\subsubsection{Combination Order}
We compared the effects of different combination orders on the performance of the results.

\begin{table}[!htp]
    \centering
    \footnotesize
    \renewcommand{\arraystretch}{1.1} 
    \setlength{\tabcolsep}{4pt} 
    \begin{tabular}{lccc}\toprule
    \textbf{Order} & \textbf{QNLI} & \textbf{SST2} & \textbf{MMLU} \\ \midrule
    Para + BT & \textbf{10.67} & \textbf{6.00} & \textbf{14.67}  \\
    BT + Para & 10.67 & 6.00 & 12.67  \\
    \bottomrule\end{tabular}
    \caption{Performance gap with different combination orders of paraphrase and back-translation.}
    \label{tab:combing_order}
\end{table}

\paragraph{Results.} From Table \ref{tab:combing_order}, we can see that while QNLI and SST2 tasks are less sensitive to method order, the MMLU task shows slight differences. Therefore, we can tailor the order based on task requirements, and we choose first to paraphrase and then back-translation in \textit{Clean-Eval}.

\subsubsection{Equivalence Detector}
Continuous back translation would end up with a string that differs markedly from that which you started \cite{way2013emerging}. A combination of paraphrase and back-translation might also cause this problem. 

\begin{table}[!htp]
    \centering
    \footnotesize
    \renewcommand{\arraystretch}{1.1} 
    \setlength{\tabcolsep}{5pt}
    \begin{tabular}{lccc}\toprule
    \textbf{Method} & \textbf{QNLI} & \textbf{SST2} & \textbf{MMLU}\\ \midrule
    BT & 74.17 & 86.33 & 73.33 \\
    Para & 91.67 & 82.67 & 73.33 \\
    Clean-Eval (w/o detector) & 72.17 \textcolor{red}{$\downarrow$} & 56.34\textcolor{red}{$\downarrow$} & 60.34\textcolor{red}{$\downarrow$} \\
    \bottomrule\end{tabular}
    \caption{Model performance on the calibrated dataset without equivalence detector.}
    \label{tab: detector}
\end{table}

\paragraph{Results.} As we can see from Table \ref{tab: detector}, across all three datasets, the paraphrasing method demonstrates relatively high performance, especially in QNLI and SST2.  Without a semantic detector, results generated through \textit{Clean-Eval} exhibit a general decline in performance. This suggests the possibility of introducing semantic errors or inaccuracies during the generation process and the importance of semantic detectors.

\subsection{Human Evaluation}
We performed human evaluations of the generated output to assess potential changes after our method \textit{Clean-Eval}. 

\paragraph{Results.} Human evaluation results on the SST2 dataset indicate that $97\%$ of instances maintain semantic equivalence with the original ones. This suggests the \textit{Clean-Eval} largely preserves the original data's intended meaning, showcasing the effectiveness in retaining input semantics.

%% file: case_study.tex
\section{Case Study}
\begin{figure}[!htp]
    \centering
    \includegraphics[width=7.5cm]{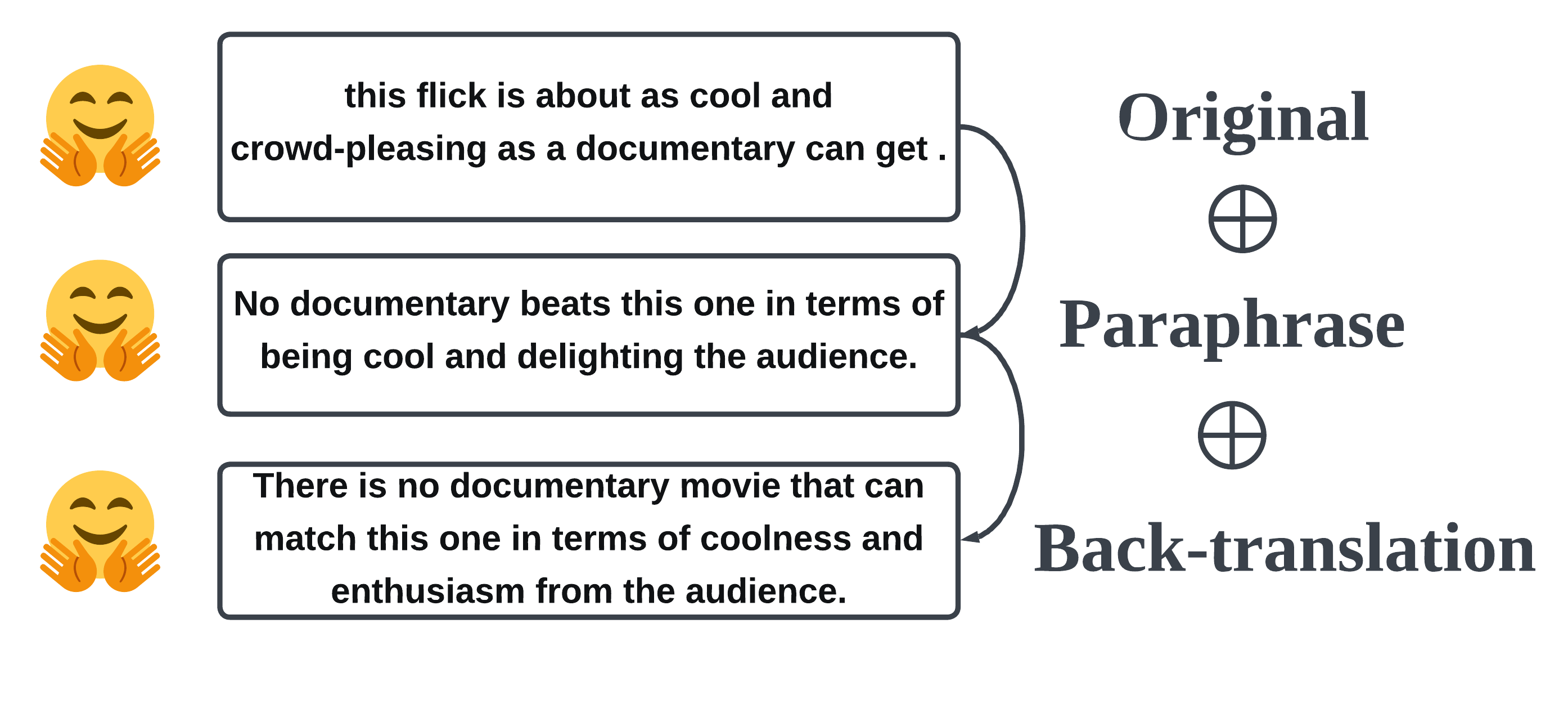}
    \caption{A case study from SST2 dataset.}
    \label{fig:case_study}
\end{figure}

In this case, the paraphrased sentence successfully conveys the essence of the original while introducing some variation. The transformation maintains a positive sentiment, emphasizing the documentary's coolness and appeal to the audience. Back-translation aims to ensure that the paraphrased sentence retains its intended meaning. The back-translated sentence aligns well with the paraphrased version. The key elements, such as the documentary's uniqueness, coolness, and audience appeal, are preserved. The combined approach of paraphrasing and back-translation effectively enhances the original sentence. The paraphrased version introduces a nuanced expression, and the subsequent back-translation successfully captures the intended meaning. The final output maintains a positive tone and successfully communicates the documentary's appeal.

%% file: conclusion.tex
\section{Conclusion}

Data contamination is an urgent problem for the development of LLMs society. Downloading and trying contaminated models can be a waste of time for both researchers and developers. To save their time, this paper intends to mitigate the issue of data contamination in LLMs by introducing the \textit{Clean-Eval} method. This approach leverages existing datasets to create a new evaluation dataset, effectively mitigating the impact of contamination. Experimental results demonstrate the method's success in accurately assessing model capabilities. \textit{Clean-Eval} holds promise in enhancing transparency and reliability in evaluating LLMs. Future work can be dedicated to co-training a data contamination detector in a neural-based framework with \textit{Clean-Eval} in a multi-tasking fashion. Additionally, we hope to open-source various versions of intentionally contaminated LLMs and their contamination information for research purposes.